# Explainable Multi-Label Classification of MBTI Types

Siana Kong

EECS, University of Ottawa
skong005@uottawa.ca

Marina Sokolova

EECS and SEPH, University of Ottawa
sokolova@uottawa.ca

**Abstract**

In this study, we aim to identify the most effective machine learning model for accurately classifying Myers-Briggs Type Indicator (MBTI) types from Reddit posts and a Kaggle data set. We apply multi-label classification using the Binary Relevance method. We use Explainable Artificial Intelligence (XAI) approach to highlight the transparency and understandability of the process and result. To achieve this, we experiment with glass-box learning models, i.e. models designed for simplicity, transparency, and interpretability. We selected k-Nearest Neighbours, Multinomial Naive Bayes, and Logistic Regression for the glass-box models. We show that Multinomial Naïve Bayes and k-Nearest Neighbour perform better if classes with Observer (S) traits are excluded, whereas Logistic Regression obtains its best results when all classes have > 550 entries.

# 1. Introduction

An individual's personality refers to their unique set of traits and behaviour, as well as the patterns of thinking and feeling (Cherry, 2023). Although everyone's personality is unique to themselves, there exist various frameworks and models to categorize personalities into distinct types. By identifying common patterns of traits, behaviors, and characteristics, these tools categorize personalities to group people with similar personalities together. This categorization can help people to understand themselves better, as well as other people by learning about different personality types. Myers-Briggs Type Indicator (MBTI) is an indicator widely used today. MBTI was developed by Katharine Cook Briggs and Isabel Briggs Myers, inspired by Carl Jung's theory of psychological types. It assigns personalities into Energy [Extraverted(E) / Introverted(I)], Mind [Intuitive(N) / Observant(S)], Nature [Feeling(F) / Thinking(T)], and Tactics [Judging(J) / Prospecting(P)], where the different combinations of these four letters make up different personality types (NERIS Analytics Limited, 2013).

We hypothesize that people with similar MBTI traits are likely to show similar behavior and interests online. Therefore, our main assumption is that MBTI type of individuals can be predicted based on the posts they make online. To test the hypothesis, we decided to experiment with different machine learning models to find the most fitting model for predicting a person's MBTI type based on the posts' text.

We use the posts from each different MBTI type's subreddits on Reddit to show that people's personality type influences the way they behave and the topics they discuss in the online world. To make the results easier to understand and interpret, we employ the Explainable Artificial Intelligence (XAI) approach. XAI puts focus on "producing more explainable models, while maintaining a high level of learning performance" (Turek, n.d.). For prospective end users of AI applications and Machine Learning algorithms, XAI helps to understand the results better, enabling to analyze and compare different outcomes more easily, emphasizes the transparency of the process, which allows us to understand how the results came to be without having to dissect the entire process.

We use glass-box models, one of the most straightforward approaches of XAI. Those are models that are easy to follow, and their decision making processes are straightforward and easy to understand, the key aspect of XAI (Roelenga, 2021). The key advantages of Glass-box models include:
- ➢ Transparency: able to observe exactly how the inputs results in such outputs
- ➢ Simplicity: use of straightforward methods that are easy to follow
- ➢ Interpretability: easy to interpret why and how certain decisions are made

For our learning, we selected Multinomial Naive Bayes, k-Nearest Neighbour (k-NN), and Logistic Regression.

The Glass-box nature of Multinomial Naive Bayes comes from its simple probability-based decision making process and the visibility of feature importance. The probabilities are directly derived from the data with no complex additional calculations, and each feature contributes independently to the probability of each class label, meaning that we can easily identify the importance for each feature.

The Glass-box nature of k-NN comes from its simple decision making rule, as well as its adjustability and interactivity. As users, we can easily adjust the k parameter to see immediately how the model's predictions change, which allows us to easily interpret the feasibility of the model by experimenting with different k values.

The Glass-box nature of Logistic Regression comes from its interpretability. It is straightforward and easy-to-understand as it uses a simple formula as explained above to predict the probability between two possible outcomes. These probabilities can be directly interpreted, allowing us to see how changing the parameters and feature values are associated with changes in the probability with different outcomes.

On the Reddit data set, Multinomial Naive Bayes and k-Nearest Neighbour obtained better results when excluding classes with S traits whereas Logistic Regression gave the best results when excluding classes with less than 550 entries The Exact Match results are 0.597, 0.515, 0.586 respectively, Hamming Loss – 0.150, 0.201, 0.178, and Macro F-score – 0.856, 0.815, 0.833.

## 2. Myers-Briggs Type Indicator

Myers-Briggs Type Indicator (MBTI) is one of the most popular personality type assessment methods that analyze a person's personality as one of 16 types. According to NERIS Analytics Limited (2013), each MBTI type is a unique combination of four different core traits and each trait is classified into two different classes.

1. Energy [Extraverted(E) / Introverted(I)] - "how one interacts with their surroundings"
2. Mind [Intuitive(N) / Observant(S)] - "how one sees the world and processes information"
3. Nature [Feeling(F) / Thinking(T)] - "how one makes decisions and copes with emotions"
4. Tactics [Judging(J) / Prospecting(P)] - "one's approach to work, planning, and decision-making"

For example, a person who is extraverted, intuitive, feeling, and judging, would have an MBTI type of ENFJ. With all the possible combinations of the 4 traits we decided on, we have the following 16 MBTI types; ENFJ, ENFP, ESFJ, ESFP, ENTJ, ENTP, ESTJ, ESTP, INFJ, INFP, ISFJ, ISFP, INTJ, INTP, ISTJ, and ISTP. Each MBTI type represents a different personality, and different combinations of each trait are separated into two different layers; the inner layer - Roles, and the outer layer - Strategies (NERIS Analytics Limited, 2013).

For this study, we only focus on the Role layer. The Role layer gives us insights into the personality type's behaviours and interests. There are four different roles: Analyst, Diplomat, Sentinel, and Explorer, where each role indicates a different combination of two different traits. *Analysts* are MBTI types with NT traits (ENTP, ENTJ, INTP, INTJ). Analysts are driven by rationality and independence. They are often considered strategic thinkers but may have more difficulties with social or romantic situations due to their utilitarian approach and strong-willed nature. *Diplomats* are the types with NF traits (ENFP, ENFJ, INFP, INFJ). Their main characteristics are empathy and cooperation. They are imaginative yet strive for harmony, which sometimes makes it harder for them to make cold rational, or difficult decisions. *Sentinels* are the SJ combination types (ESFJ, ESTJ, ISFJ, ISTJ). They prioritize cooperation and practicality, seeking to establish order and stability. Sentinels excel in environments with clear hierarchies and rules, which indicates that they may struggle to be flexible and appear to be stubborn when presented with different points of view. *Explorers* is the group of MBTI types with SP traits (ESFP, ESTP, ISFP, ISTP). These are the most spontaneous and adventurous types. They are practical yet flexible, which allows them to react and think quickly to unexpected situations, but this combination can also push them to take risky behaviours as they are drawn to sensual pleasures.

The main goal of this study is to show that people's personality types reflect on their posts on social media. We aim to demonstrate this by discovering the most suitable machine learning model that can predict MBTI types based on the posted texts. Findings from our work can be further used to study how MBTI type affects a person's behaviour online. Understanding these behavoiurs would greatly help with engineering a more personalized online experience by determining the user's personality type based on their posts.

## 3. Related work

There have been multiple studies done for predicting MBTI using machine learning models. To ensure that our study is unique, we avoided using the same combination of methods and models that the previous studies have already experimented with. At the same time, we use the previous results as a comparison.

Hernandez and Knight (2017) experiment with various types of Recurrent Neural Network (RNN) to predict the MBTI type. They concluded that the LSTM option gave the best results out when compared with SimpleRNN, GRU, and Bidirectional LSTM. Hernandez and Knight tested their model with Donald Trump's tweets on Twitter. Their model successfully predicted Donald Trump's MBTI correctly, which shows the real-life application of the prediction model. Hernandez and Knight's model produced the average of the probabilities for each label. For our study, we used different sets of classification models, namely kNN, Logistic Regression, and Multinomial Naive Bayes, with a binary relevance method. Using the binary relevance method, our models did binary classification on each label, instead of producing the average of probabilities for each label.

Zhang (2023) uses two word-embedding methods, BERT classification and TF-IDF Vectorizor, and three models, Logistic Regression, K-Nearest Neighbours, and Random Forest Classifier, to find the best combination for predicting MBTI type. The results showed that BERT classification with Logistic Regression gave the best results with an average accuracy of 87%. For our study, we made similar choices for the models, using Logistic Regression and kNN as well as our glass-box models for the XAI approach. We, however, used TF-IDF Vectorizor to obtain the best results.

Kaushal et al. (2021) use sentiment analysis on top of MBTI prediction to evaluate the person's sentiment about specific topics. They did not directly combine the sentiment analysis and MBTI type prediction together but rather did two separate sets of experiments. The goal of their study is to combine the advantages of the two experiments to be used in real-life applications such as in the hiring system for selecting candidates that are most suitable for their company based on the candidates' social media posts. They conducted experiments for MBTI type prediction using K-Nearest Neighbours, Logistic Regression, Naive Bayes, Random Forest, SVM, and Stochastic Gradient Descent and concluded that SVM outperformed other algorithms. Using tweets from Twitter as their dataset, their results showed the best accuracy for the NS label. In our work, we focus on multi-label classification of MBTI types.

## 4. The MBTI Datasets

In this study, we used two datasets from two different sources; Reddit and Kaggle. We applied the same pre-processing steps to both datasets: Convert all letters to lowercase, Remove emoticons, Remove URLs, Remove punctuation, Tokenization, Remove Stop words, Lemmatization, Remove numbers by themselves

### 4.1 The Reddit data set

W collected the Reddit data by extracting up to 1000 most recent posts from each MBTI type (https://www.reddit.com/r/mbti). We selected 1000 posts as the limit to prevent a big imbalance between the types as some MBTI subreddits did not have 1000 posts but were close to it. We combined all the data extracted from the subreddit together with the post and the MBTI type it belongs to based on the subreddit the post was extracted from. However, the MBTI type of some posts had to be adjusted manually because some people would post on MBTI subreddits that do not match their own MBTI type. For example, a person of INTP MBTI type would have a post on the ESFJ subreddit to ask questions about the ESFJ type. We deleted the posts where the author's MBTI type is unclear to reduce any confusion for the training models.

After the preprocessing step, 14,582 rows of data remained. We then decided on the minimum and maximum number of tokens range by getting the outlier range for the minimum and maximum number of tokens first and then applying the Type-to-Token Ratio (TTR) to different ranges. Calculating the minimum and maximum token ranges helps to eliminate posts that are too short or too long and are considered the outlier posts. This step enhances the data quality by defining a normal data point, in this case, the number of tokens in posts. It can also improve model accuracy by training the model more effectively on the core data, i.e. the posts that fall under the range. The minimum number of tokens for the outlier range came down to a negative number, -82, which means there is no outlier for the lower bound. The maximum number of tokens for the outlier range came down to 166, meaning that any post data with more than 166 tokens is to be considered an outlier and, therefore, eliminated. These outlier bounds were calculated using the $1.5 \times$ Inter-Quartile Range (IQR) rule, where we set the $k$ value to 1.5 in the following equations for defining the lower and upper outlier boundaries.

Calculating Inter-Quartile Range (IQR):

$$IQR = Q_3 - Q_1$$
$$Q_1 = \text{first quartile (25th percentile)}$$
$$Q_3 = \text{third quartile (75th percentile)}$$

Defining the outlier boundaries:

$$\text{Lower bound} = Q_1 - k \times IQR$$
$$\text{Upper bound} = Q_3 + k \times IQR$$

The TTR formula calculates the ratio of the number of unique words to the total number of words in the dataset. A higher TTR score is preferred as it indicates richer vocabulary, which improves model performance as it becomes easier to interpret more variety of nuances in the text. This means that experimenting with a dataset with higher TTR would train the model with more diversified data, dealing with fewer posts with the same set of vocabulary.

$$\text{Type-to-Token Ratio} = \frac{\text{Total number of unique words}}{\text{Total number of words}}$$

After experimenting with different ranges within the outlier range (0-166 tokens), we set the minimum number of tokens to be 11, and the maximum to be 166. We decided on this range because this range had the highest TTR while also reserving 10,000+ data. We noticed that if we set the minimum range to higher than 11, it returns a higher TTR but returns less than 10,000 rows of data. The TTR after these boundaries are applied to the Reddit dataset returned 0.304. Table 1 shows the number of data in the final Reddit dataset for each MBTI type after preprocessing and manual adjustments have been made, with a total of 10,086 rows of data:

Table 1: Each MBTI type and the number of examples for the Reddit data

| type | # | type | # | type | # | type | # |
|------|-----|------|-----|------|-----|------|-----|
| ENFJ | 712 | ESFJ | 565 | INFJ | 830 | ISFJ | 409 |
| ENFP | 759 | ESFP | 550 | INTP | 597 | ISFP | 588 |
| ENTJ | 779 | ESTJ | 490 | INTJ | 749 | ISTJ | 554 |
| ENTP | 703 | ESTP | 518 | INTP | 767 | ISTP | 516 |

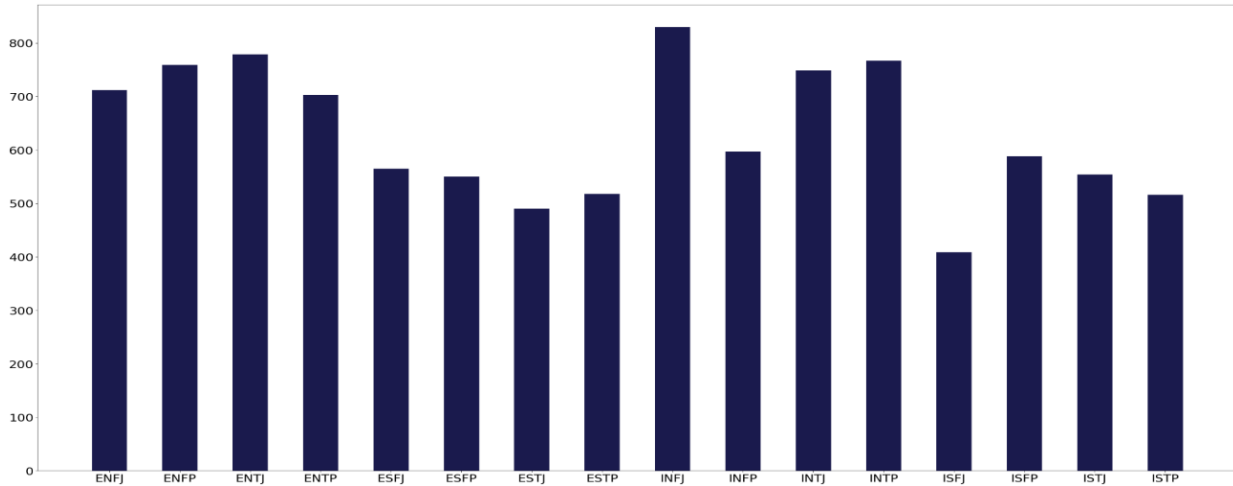

Figure 1. Distribution of the 16 MBTI types for Reddit data

There appears to be a slight imbalance between a few MBTI types, visualized by Fig. 1, where most types with Intuitive trait (N) have more than 700+ data and the types with Observant trait (S) have less than 600 data, which are the two opposing classes of the second trait, the Mind, of MBTI type. We speculate that this pattern may be due to the N (Intuitive) class being described as the "imaginative, open-minded, and curious" class. Therefore, it is possible that the people with the N trait make more posts online as they may have more questions or curious topics they want to talk about compared to the opposing S (Observant) class, which is described as more "practical, pragmatic and down-to-earth".

### 4.2 The Kaggle data set

As our second dataset, we used a Kaggle dataset that already had data collected from an online forum called Personality Cafe, collected by Mitchell J, (https://www.kaggle.com/datasets/datasnaek/mbti-type). Each row of data has 50 posts made by one person along with their MBTI type, therefore, there was no need to manually fix the MBTI type for each row for the Kaggle dataset. We decided to keep the 50 posts in one line, rather than separating them into 50 rows of separate data, as they were written by the same author anyway. While manually going through the posts, we found that some posts ended with '...' without properly ending the sentence. We noticed that these posts were cut off due to the size of the post. We decided to exclude these posts because a post that is not complete can appear ambiguous and confuse the classification models when training. Instead of conducting the IQR calculation all over again for the Kaggle dataset, we decided to use the same minimum and maximum outlier boundaries, 11 and 166 words, for the Kaggle dataset as well. This turned out to be a reasonable boundary for the Kaggle dataset , as we found that the TTR, after applying the same boundaries, is very close to the TTR of the Reddit dataset; Kaggle dataset TTR: 0.298. Table 2 shows the number of data in the final Kaggle dataset for each MBTI type after preprocessing and the outlier boundaries applied, with a total of 3,886 rows of data.

Table 2: Each MBTI type and the number of examples for the Kaggle data

| type | # | type | # | type | # | type | # |
|------|---|------|---|------|---|------|---|
| ENFJ | 101 | ESFJ | 20 | INFJ | 756 | ISFJ | 81 |
| ENFP | 289 | ESFP | 19 | INTP | 881 | ISFP | 107 |
| ENTJ | 89 | ESTJ | 14 | INTJ | 515 | ISTJ | 88 |
| ENTP | 242 | ESTP | 26 | INTP | 543 | ISTP | 115 |

There is a significant imbalance between the MBTI types (Fig. 2). The types with IN combination have the most data available, and the ES combination with the least amount of data. As in the Reddit dataset, there is sufficiently more data for the types with the N trait versus the S trait, although the imbalance here is much bigger. Hence, we opt to use the Reddit data as the main dataset for our experiments because it has much more data available and less imbalance between different MBTI types. The Kaggle dataset is used for comparison to the Reddit data experiments, to see if there are any significant differences or patterns to be found.

Figure 2. Distribution of the 16 MBTI types for Kaggle data

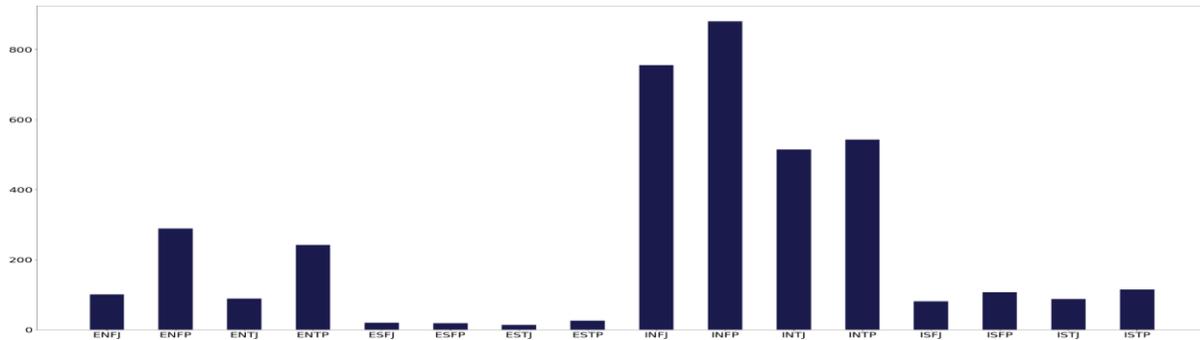

## 5. Classification Methods

Our classification method is considered a multi-label classification where each trait of the MBTI type is considered as a single label with binary classes. With multi-label classification, we are looking at each trait of the MBTI type as a label as such; Label 0: [E/I], Label 1: [N/S], Label 2: [F/T], Label 3: [J/P]. We decided that Binary Relevance would be the most appropriate method for this case. With Binary Relevance, each four traits of the MBTI type is considered as a single label, performing binary classification separately for each label (Nooney, 2018). The results are the union of all the labels' classification results.

We decided to use the following glass-box models as our classifiers; Multinomial Naive Bayes, k-Nearest Neighbours, and Logistic Regression. For all experiments, we used the 5-fold cross-validation method, which splits the dataset into 5 folds and trains and evaluates the model 5 times where one fold is used as the test set and the rest as a training set for each iteration. Then the average of the performance metrics obtained across all folds is reported as the final model performance.   Each classifier followed the following general steps with the Reddit dataset:
1. Vectorization of the text data using TF-IDF Vectorizer
2. Training and testing data using 5 fold cross validation strategy; using the full dataset
3. Repeat 2 excluding all classes with the S trait
4. Repeat 2 excluding the N/S label
5. Repeat 2 excluding classes with less than 550 entries
6. Repeat 2, 3, 4, and 5 with feature selection using SelectKBest + Chi-squared (Chi2)

The TF-IDF Vectorizer is a feature extraction technique that transforms the preprocessed tokens into numerical vectors, where each dimension represents a unique word or token, and the value of each dimension represents the TF-IDF score of the specific token. TF-IDF score is a product of TF and IDF; Term Frequency (TF) counts the frequency of each token in the post, and Inverse Document Frequency (IDF) identifies the significance of each token in the entire corpus by giving less importance to words that appear more frequently across the corpus.

We used the result of Step 2 as the baseline result for each classifier experiment, which uses all the data in the dataset. Steps 3, 4, and 5 were tested to see if minimizing the class imbalance in different ways can improve the results. For Step 3, we excluded the classes with the S trait because as explained in the Dataset section, MBTI types with the N trait tend to have 200-300 more data per type compared to the S traits. Therefore, we hypothesize that getting rid of the MBTI types with the S trait would improve the results since the N/S label will always test and predict correctly since only the N trait is left. However, this means that the Precision, Recall, and F1 Score for the N/S label will always be 1.0 with the exclusion since only data with the N class out of the two N and S classes will be used. Step 4 excludes the N/S label as a whole, to see if getting rid of this imbalanced label as a whole would improve the score of the remaining labels. Step 5 excludes MBTI types with less than 550. This way, we still get 50% of the (4 out of 8) MBTI types with the S trait but with less class imbalance. Finally, for Step 6, we repeat everything from Steps 2 to 5 but apply feature selection using SelectKBest and Chi2 (Goswami, 2020). SelectKBest is a feature selection technique that selects the top k features that are most relevant to the current class. We paired this technique with Chi2, which is a scoring method that scores categorical features based on their relationship with the current class. A higher Chi2 value indicates a greater deviation between the observed and expected frequencies, which suggests a stronger association between the variables. With the combination of SelectKBest and Chi2, we score each feature independently by their significant association with the class and select the top k features with the highest scoring. Once the top k features are selected, the dataset is transformed to include only those selected features and exclude the rest. This can improve the efficiency of the classifiers as it reduces the dimensionality of the dataset to focus on only the selected features.

Data sets used for each step:
- ➢ **Step 2:** All data in the dataset, including all MBTI classes (types); ENFJ, ENFP, ENTJ, ENTP, ESFJ, ESFP, ESTJ, ESTP, INFJ, INFP, INTJ, INTP, ISFJ, ISFP, ISTJ, ISTP
- ➢ **Step 3:** MBTI classes with N trait, excluding all classes with S trait; ENFJ, ENFP, ENTJ, ENTP, INFJ, INFP, INTJ, INTP
- ➢ **Step 4:** All MBTI classes excluding the N/S label; EFJ, EFP, ETJ, ETP, IFJ, IFP, ITJ, ITP
- ➢ **Step 5:** MBTI classes with 550+ data: ENFJ, ENFP, ENTJ, ENTP, ESFJ, ESFP, INFJ, INFP, INTJ, INTP, ISFP, ISTJ

For the experiment results, we decided to look at both label-wise classification scores and class-wise classification scores to compare the scores for both and see if there are any significant differences between them. While the label-wise results show the scores for each four labels across the entire corpus, the class-wise results show the score for each four labels for each MBTI type/ class. Therefore, the class-wise results do not indicate multi-class classification, as it still uses the multi-label classification method, but shows the results in each different combination of the four labels. For the evaluation metrics, we decided to use Exact Match Ratio, Hamming Loss, Precision, Recall, and F1 scores, and get the micro & macro averages for Precision, Recall, and F1 scores. We chose the Exact Match Ratio as it is a measure of accuracy for multi-label prediction. We chose Hamming Loss because it accounts for label imbalance by treating each label independently. We chose Precision, Recall, and F1 scores as they are commonly used metrics in many classification tasks, which we can use to compare the results between label-wise classification and class-wise classification.

## 6. Classification Results
In this section, we summarize the experiment results for each classification model we chose and compare the best results from each model. For each summary, we indicate the step we are on, as specified in Section 5, to identify which set of data is being used.

### 6.1 Multinomial Naive Bayes
The Multinomial Naive Bayes algorithm is a probabilistic classifier, which calculates the probability of a certain outcome given some features (Great Learning Team, 2024). It takes in vectorized data, TF-IDF vectorization in our case, and calculates the probability of observing each feature for each class using the training data. Using these probabilities, it computes the probability that the current post belongs to each class, and predicts the class label for the post based on the class with the highest probability. In the case of MBTI multi-label classification, it calculates the probability of the author's post belonging to a specific trait for each label. Multinomial Naive Bayes is primarily used for text classification tasks where the input data is represented as vectors, and it is particularly well-suited for tasks involving text data (Table 3).

Table 3. (Step 2) Label-wise Baseline results for Multinomial Naive Bayes

| Label-wise results | | | |
|---|---|---|---|
| Exact Match Ratio | 0.312 | Hamming Loss | 0.270 |
|  | Precision | Recall | F1-score |
| Label 0 [E/I] | 0.775 | 0.530 | 0.751 |
| Label 1 [N/S] | 0.639 | 0.989 | 0.777 |
| Label 2 [F/T] | 0.724 | 0.831 | 0.773 |
| Label 3 [J/P] | 0.697 | 0.846 | 0.764 |
| Micro average | 0.688 | 0.813 | 0.716 |
| Macro average | 0.709 | 0.849 | 0.766 |

The class-wise results were very close to the label-wise results, with the average Exact Match Ratio and average Hamming Loss of all classes being the same as the Label-wise results when reduced down to 3 decimal points. We noticed that when looking at the scores for each of the classes, the scores were very similar even for the classes with fewer data - such as the ones with S trait. We assume this is due to the fact that we are still conducting a multi-label classification and not a multi-class classification, so the training is based on the labels rather than the classes.

(Step 3) When we conducted the experiment with all the classes with the S trait excluded, the exact match ratio, hamming loss, and micro & macro average scores improved quite a bit compared to the baseline results. The exact match ratio almost doubled and the hamming loss went down by 0.07. However, the per-label scores did not improve for the remaining labels did not improve. We predict that this is due to the fact that Label 1 would always be predicted correctly since there exists only one class (N) for the binary classification of the label.

(Step 4) Next, we excluded the [N/S] label itself, to see if getting rid of the imbalanced label entirely would improve the overall score as well. The results showed a higher exact match ratio and lower hamming loss compared to the baseline results. However, micro and macro averages of precision, recall, and f1-score scored lower compared to the baseline results. The per-label scores also stayed exactly the same as the baseline results, which indicates that excluding the [N/S] label did not affect the other labels. This is as predicted since we are using Multinomial Naive Bayes, which assumes independence between features.

(Step 5) When we excluded all classes with less than 550 data, the exact match ratio got lower and all the metrics were mostly similar compared to the baseline results. However, the recall rates were noticeably higher, which indicates that this dataset minimizes false negatives.

(Step 6) After trying different combinations of data, we went on to apply feature selection to the different combinations. We decided to only use the MBTI types excluding classes with S traits because the other two combinations' results did not show any improvement from the baseline results. With the SelectKBest + Chi2 combination for feature selection, we experimented with different values of k for SelectKBest, starting from 50 and incrementing until we found the best k value. When using the entire dataset with no restrictions, we found the best result with k_best = 150 when experimented up to k_best = 250. With k_best = 150, the exact match ratio and per-label precision scores improved, but the micro averages of precision, recall, and f1-score were all slightly lower compared to the results of the experiment using a dataset excluding classes with S trait with no feature selection.

**Table 4. (Step 6) Best label-wise results for Multinomial Naive Bayes**

| Label-wise results |  |  |  |
|---|---|---|---|
| Exact Match Ratio | **0.597** | Hamming Loss | **0.150** |
|  | Precision | Recall | F1-score |
| Label 0 [E/I] | **0.826** | **0.769** | **0.795** |
| Label 1 [N/S] | **1.000** | **1.000** | **1.000** |
| Label 2 [F/T] | **0.830** | **0.803** | **0.814** |
| Label 3 [J/P] | **0.720** | **0.936** | **0.813** |
| Micro averages | **0.873** | **0.911** | **0.873** |
| Macro average | **0.844** | **0.877** | **0.856** |

We achieved the best results with Multinomial Naive Bayes when we combined feature selection with the S trait excluded dataset, with k_best = 250, when tested up to k_best = 300. With this combination, we obtained the best results out of all experiments we have conducted for multinomial naive bayes (Table 4)

## 6.2 k-Nearest Neighbours

K-Nearest Neighbours (kNN) aims to predict the class of a test data point by comparing its distance to all training data points (Commercial Data Mining, 2022). It then identifies the k nearest neighbours to the test point based on the previously calculated distance. These nearest neighbours are then used to predict the class for the current test data point. For deciding on the k value, we started by following the general rule of thumb for choosing the k value, which is setting it to the square root value of the size of the training data. Since we had about 8000 samples in the training data, we initially ran the experiments with 89 (sqrt of 8000), which gave us the result found in Table 4. Then we also checked the results for setting the k to number lower than and greater than 89, using 87 and 91, and found that 89 still gives the best results. We also tried with much lower k values starting from 1, however, the overall results were much worse compared to when k is set to 87, 89, or 91.

Table 5. (Step 2) Label-wise Baseline results for kNN

| Label-wise results | | | |
|---|---|---|---|
| Exact Match Ratio | 0.391 | Hamming Loss | 0.265 |
| | Precision | Recall | F1-score |
| Label 0 [E/I] | 0.770 | 0.663 | 0.712 |
| Label 1 [N/S] | 0.728 | 0.943 | 0.823 |
| Label 2 [F/T] | 0.682 | 0.845 | 0.754 |
| Label 3 [J/P] | 0.698 | 0.792 | 0.742 |
| Micro average | 0.708 | 0.777 | 0.714 |
| Macro average | 0.719 | 0.811 | 0.758 |

When looking at the class-wise results (Table 5), it follow the same pattern as the Multinomial Naive Bayes, where the class-wise and label-wise results appear to be very similar. The overall scores for the baseline results for both kNN and Multinomial Naive Bayes appear close to each other.

(Step 3) When conducting the experiment excluding all classes with the S trait, we saw an overall improvement in all scores for both label-wise and class-wise results. The exact match ratio increased to 0.478 and hamming loss went down to 0.196, and the micro & macro average scores increased by about 0.1 for each metric. Unlike Multinomial Naive Bayes, the per-label score was affected by the absence of the S trait classes when using kNN, because kNN directly relies on the neighbouring points in feature space. Therefore, it makes sense that an absence of a label would affect the distances and relationships between data points, potentially leading to changes in prediction values for the other labels. However, although the per-label score was affected, the change in the scores itself appeared to be very minimal.

(Step 4, 5) We were not able to conduct experiments for excluding the [N/S] label and excluding classes with less than 550 data for kNN due to hardware problems, as we kept running out of storage space when trying to run these experiments.

(Step 6) We only conducted the feature selection application with all data combined and classes with the S trait excluded, as we could not run the experiments for the other two cases. We used the same method for selecting the k best value for SelectKBest as we did with Multinomial Naive, Bayes. When conducting the experiment with all data combined, we found the best results using k_best = 50, after experimenting with k_best = 50, 100, and 150, as the scores gradually went down as the k value increased. For the metrics scores, the exact match ratio improved to 0.489, but the score for all the other metrics stayed almost the same as the baseline result. This result proves to show that kNN with k = 89 has strong fidelity because the results did not change when feature selection was applied. Strong fidelity is important as it indicates the classifier's ability to be consistent in learning. When experimented with a much lower k value, such as k = 5, the exact match ratio appeared to be higher than when k = 89. However, with k = 5, all the metrics changed drastically when comparing results between with or without the feature selection applied, meaning that lower k numbers, such as k = 5, are unstable compared to k = 89.

**Table 6. (Step 6) Best label-wise results for kNN**

| Label-wise results | | | |
|---|---|---|---|
| Exact Match Ratio | **0.515** | Hamming Loss | **0.201** |
| | Precision | Recall | F1-score |
| Label 0 [E/I] | **0.704** | **0.742** | **0.722** |
| Label 1 [N/S] | **1.000** | **1.000** | **1.000** |
| Label 2 [F/T] | **0.682** | **0.876** | **0.767** |
| Label 3 [J/P] | **0.715** | **0.840** | **0.772** |
| Micro average | **0.845** | **0.904** | **0.841** |
| Macro average | **0.775** | **0.864** | **0.815** |

When conducting the same experiment using data excluding all classes with the S trait, we found a slight improvement in exact match ratio, hamming loss, and micro average scores. However, the macro average and per-label scores remained about the same. The increase in micro averages while the macro average remaining the same suggests that the classifier is getting better at predicting the correct outcome for each data point overall, without favoring specific classes over others. This once again proves the consistency of this classifier, indicating strong fidelity. Table 6 shows the results for feature selection applied to data excluding all classes with S trait, with k = 89 and k_best = 50, which provides the best results for experiments with kNN.

## 6.3 Logistic Regression

Logistic Regression works by drawing a line, that works as a decision boundary, to separate data points that represent two different groups (Commercial Data Mining, 2014). One side of the line represents group A, and the other side of the line represents group B. The goal of this classifier is to find the best line that minimizes the error in categorizing points into the correct groups, and once this line is determined, it is used to predict which group the new data point belongs to based on its location relative to the line.

Table 7. (Step 2) Label-wise Baseline results for Logistic Regression

| Label-wise results | | | |
|---|---|---|---|
| Exact Match Ratio | 0.516 | Hamming Loss | 0.201 |
| | Precision | Recall | F1-score |
| Label 0 [E/I] | 0.824 | 0.744 | 0.782 |
| Label 1 [N/S] | 0.790 | 0.942 | 0.859 |
| Label 2 [F/T] | 0.801 | 0.787 | 0.794 |
| Label 3 [J/P] | 0.799 | 0.771 | 0.785 |
| Micro average | 0.770 | 0.776 | 0.750 |
| Macro average | 0.804 | 0.811 | 0.805 |

While the overall scores for the baseline results for kNN and Multinomial Naive Bayes appear close to each other, the baseline results for Logistic Regression appear to start off with much improved scores from the start compared to the other two classifiers (Table 7).

For Logistic Regression, we experimented with different combinations of the following parameters as well; C, Solver, Penalty, and Max_iter. We tried every combination of different values for each parameter and found that it returns the best results when Solver = liblinear,

Penalty = l1, C to its default value which equals to 1, and Max_iter also to its default value, 100, because changing the Max_iter did not affect the results at all. With these parameters applied to the baseline experiment, the results showed slight improvement for an exact match ratio of 0.584 and hamming loss of 0.186, but the other metrics scores did not improve. In fact, the per-label metrics scores slightly decreased. This implies that the overall accuracy of predicting the correct combination of labels for each instance has increased, but the classifier is making more mistakes in correctly predicting the specific labels for each instance. This may be due to the parameter tuning causing a trade-off between optimizing for overall accuracy and per-label precision.

(Step 3) We could not conduct an experiment with data excluding all classes with the S trait for Logistic Regression. While the other classifiers allowed for the absence of the S trait for the [N/S] label classification, sklearn's Logistic Regression tool required at least 2 classes in the data, as this algorithm works by finding the relationship between two different groups of data. Thus, the complete absence of the S trait arose as a problem for the classification of the [N/S] label. Logistic Regression cannot learn anything meaningful with only one class present as it tries to find the decision boundary that best separates the classes. Therefore, with only one class, there is no decision boundary to learn, leading to the following error we have encountered.

However, this was not the case with kNN or Multinomial Naive Bayes. As explained above, kNN operates by finding the k nearest neighbours based on distance metrics and assigns the class based on the majority class among the neighbours. Multinomial Naive Bayes assumes independence between features given the class. Therefore, when there's only one class available, it may still compute probabilities using the available features.

(Step 4) When we completely excluded the [N/S] label with default parameters, the exact match ratio appeared to have improved compared to the baseline results, but all the other scores were generally worse. Also, the baseline experiment with the parameters applied had a higher exact match ratio and all other scores than the experiment results of Step 4. We decided to run the experiment again with the Step 4 dataset with the same parameters applied. The results showed a much higher exact match ratio of 0.625, but the micro average of precision, recall, and F1 score has dropped lower. This means that the correct classification of exact label combinations may have improved, but the overall performance across all labels has been downgraded.

(Step 5) For the dataset excluding all MBTI types with less than 550 data, we also experimented with both the default parameters and parameters applied. With default parameters, the exact match ratio appeared slightly lower than the baseline result, with a slight improvement in micro average scores. But with the parameters applied, it gave the best overall result so far, with an exact match ratio of 0.586, along with higher precision, recall, and F1 scores.

(Step 6) Finally, we applied feature selection to the experiments conducted so far. When applying feature selection to all data combined with the parameters, we found the best results using k_best = 200, after experimenting with k_best = 50, 100, 150, 200, and 250. This experiment showed similar results to the one of Step 5, which excludes the MBTI types with less than 550 data with parameters applied, but had lower micro average scores. Similarly, when applying feature selection to data excluding the [N/S] label with the parameters with k_best = 150 (best out of k = 50, 100, 150, 200, 250), it returned a high exact match ratio of 0.628, but there appeared to be a big decrease in micro average scores. Lastly applying feature selection to data excluding MBTI types with less than 550 data with parameters applied and k_best = 150 (best out of k = 50, 100, 150, 200, 250), we saw that the results were very similar to the one of same combination without the feature selection, which has shown the best overall results so far. This result proves that like the kNN results, this set of data and parameters has strong fidelity because the results did not change when feature selection was applied.

**Table 8. (Step 5) Best label-wise results for Logistic Regression**

| Label-wise results | | | |
|---|---|---|---|
| Exact Match Ratio | **0.586** | Hamming Loss | **0.178** |
| | Precision | Recall | F1-score |
| Label 0 [E/I] | **0.896** | **0.700** | **0.785** |
| Label 1 [N/S] | **0.870** | **0.981** | **0.921** |
| Label 2 [F/T] | **0.794** | **0.889** | **0.839** |
| Label 3 [J/P] | **0.830** | **0.751** | **0.788** |
| Micro average | **0.846** | **0.853** | **0.829** |
| Macro average | **0.847** | **0.830** | **0.834** |

The best result for Logistic Regression is given by data excluding MBTI types with less than 550 data, along with the parameters Solver = liblinear and Penalty = L1 (Table 8).

### 6.4 Selecting the best learning model

In summary, Multinomial Naive Bayes showed the best results with data excluding classes with S traits with feature selection with k_best = 250, kNN also showed the best results with data excluding classes with S traits with k = 89 and features selection with k_best = 50, and Logistic Regression gave the best results with data excluding classes with less than 550 entries with solver = liblinear and penalty = L1 as its parameters. Table 9 shows the scores for each metric we decided to use to compare the best results with each classification model.

Table 9. Metrics of best results from each classification model

|  | Multinomial NB | kNN | Logistic Regression |
|---|---|---|---|
| Exact Match Ratio | 0.597 | 0.515 | 0.586 |
| Hamming Loss | 0.150 | 0.201 | 0.178 |
| (Micro) Precision | 0.872 | 0.845 | 0.846 |
| (Micro) Recall | 0.911 | 0.904 | 0.853 |
| (Micro) F1 | 0.873 | 0.841 | 0.829 |
| (Macro) Precision | 0.844 | 0.775 | 0.847 |
| (Macro) Recall | 0.877 | 0.864 | 0.830 |
| (Macro) F1 | 0.856 | 0.815 | 0.833 |

The kNN result shows the worst score overall while Multinomial NB and Logistic Regression show similar results with Multinomial NB scores being slightly better. Since the Multinomial NB result is based on excluding the classes with the S trait, it is ought to always give a full (1) score for precision, recall, and F1 scores for the N/S label, which contributes to the overall higher score of micro and macro averages of the scores. However, the Logistic Regression result is based on data excluding classes with less than 550 data, which still leaves us with 50% of the S trait classes. Therefore, although the scores of Multinomial NB best result are slightly higher than the one of Logistic Regression, we concluded that the Logistic Regression experiment with this set of data is the best model for predicting MBTI types as it is trained to predict the types with no complete exclusion of any trait. In conclusion, we decided that the following stack is the best MBTI prediction model from the experiments conducted in this study.

➢ Method: Multi-label classification using Binary Relevance
➢ Classification model: Logistic Regression (solver = liblinear, penalty = L1)
➢ Data: MBTI types with 550 or more data; ENFJ, ENFP, ENTJ, ENTP, ESFJ, ESFP, INFJ, INFP, INTJ, INTP, ISFP, ISTJ

**6.5 T-test on the Best Results**

With our best result, we conducted a t-test between each pair of different labels to identify any statistical significance (Bevans, R., 2023). We decided to conduct the t-test between each label with their precision, recall, and F1 scores. Out of each set of pairs we experimented with - [E/I] & [N/S], [E/I] & [F/T], [E/I] & [J/P], [N/S] & [F/T], [N/S] & [J/P], [F/T] & [J/P] - only the t-test with [J/P] and [N/S] labels scores showed statistical significance with a p-value of 0.027. Looking at the label-wise results of the best model selected (Tabel 8), Label 1 [N/S] has the best overall score, and Label 3 [J/P] has the lowest overall score, which we assume is the reason why the statistical significance between these two labels specifically. Since the [N/S] scores, 0.870, 0.981, and 0.921 (Precision, Recall, and F1 scores respectively) are higher than the [J/P] scores, 0.830, 0.751, and 0.788, the statistical significance implies that the [N/S] label performs better than the [J/P] label in this context, meaning that the model is more effective at correctly predicting the [N/S] label.

### 6.6 The Kaggle data experiment summary

Due to a significant imbalance in Kaggle data, we opted to use it in some experiments that were conducted with the Reddit data. We only conducted the Kaggle data experiments with the same conditions applied from the baseline and best results experiments from the Reddit data. Our goal was to identify if there is any significant difference between results on both data sets.
In our experiments, Logistic Regression was the only classification model out of the three models that was able to return the full baseline result (Table 10).

Table 10. Label-wise Baseline results for Logistic Regression

| Label-wise results | | | |
|---|---|---|---|
| Exact Match Ratio | 0.367 | Hamming Loss | 0.227 |
| | Precision | Recall | F1-score |
| Micro averages | 0.814 | 0.752 | 0.752 |
| Macro averages | 0.802 | 0.576 | 0.570 |
| Label 0 [E/I] | 0.880 | 0.024 | 0.047 |
| Label 1 [N/S] | 0.879 | 0.999 | 0.936 |
| Label 2 [F/T] | 0.724 | 0.900 | 0.802 |
| Label 3 [J/P] | 0.722 | 0.379 | 0.496 |

The exact match ratio and hamming loss scores of the baseline Logistic Regression experiment result using Kaggle data appeared to be much worse compared to the one using Reddit data. This is as expected as we have much less data and a huge class imbalance in the Kaggle data. However, the micro-average scores remained about the same. This indicates that the model is still predicting individual labels with similar accuracy, but predicting the whole set of labels correct for each type is less accurate. We also noticed that the recall and f1 scores are very low compared to the high precision score for label 1, meaning that the model has a high rate of true positives compared to false positives, but also fails to identify a significant number of actual positives. This may be due to the fact that there is much more data available for the I trait compared to the E trait within the Kaggle dataset.

Table 11 shows the results of experiments of the Kaggle dataset with the same set of data and combinations of parameters of the best results we got from using the Reddit dataset.

Table 11. Metrics of Reddit's best results applied to the Kaggle dataset

|  | Multinomial NB | kNN | Logistic Regression |
|---|---|---|---|
| Exact Match Ratio | 0.264 | 0.439 | 0.450 |
| Hamming Loss | 0.259 | 0.202 | 0.199 |
| (Micro) Precision | 0.799 | 0.849 | 0.853 |
| (Micro) Recall | 0.758 | 0.847 | 0.808 |
| (Micro) F1 | 0.752 | 0.818 | 0.801 |
| (Macro) Precision | NaN | 0.727 | 0.776 |
| (Macro) Recall | 0.505 | 0.654 | 0.631 |
| (Macro) F1 | 0.446 | 0.640 | 0.658 |

Logistic Regression showed the best results on the Kaggle dataset. However, kNN showed gave almost the same results as Logistic Regression, while there was not much improvement with the Multinomial Naive Bayes model. We assume that having such a different dataset with a significant imbalance may have caused this result, indicating that the model and the combination of parameters that gave the best results for the Reddit dataset may not be best for the Kaggle dataset. We noticed that the macro-average scores are noticeably lower compared to the micro-average scores, which was not the case with the Reddit best results experiments. This is expected due to the class imbalance we have, as this pattern indicates that the models perform well on the majority classes with more samples, which inflates the micro averages since it accounts for class size.

While the Logistic Regression experiment using the dataset excluding MBTI types with less 550 data, with Solver = liblinear and Penalty = L1, once again produced the best results out of the three models, the overall scores obtained on the Kaggle data are still worse compared to the Reddit dataset (Table 12). Per-label precision scores got noticeably worse compared to the baseline results, except for label 1. From re-running the core experiments with the Kaggle dataset and comparing the results to the Reddit dataset, we learned that having a big class imbalance and less data available has a big impact on the model's predictability.

Table 12. Best label-wise results with the Kaggle dataset

| Label-wise results | | | |
| --- | --- | --- | --- |
| Exact Match Ratio | **0.450** | Hamming Loss | **0.199** |
|  | Precision | Recall | F1-score |
| Label 0 [E/I] | **0.664** | **0.162** | **0.260** |
| Label 1 [N/S] | **0.937** | **0.998** | **0.967** |
| Label 2 [F/T] | **0.748** | **0.860** | **0.800** |
| Label 3 [J/P] | **0.755** | **0.502** | **0.603** |
| Micro averages | **0.853** | **0.808** | **0.801** |
| Macro averages | **0.776** | **0.631** | **0.658** |

## 7. Conclusions and Future Work

In this study, we experimented with kNN, Multinomial Naive Bayes, and Logistic Regression classification models using the Binary Relevance method for multi-label classification of MBTI types based on Reddit and Kaggle data. We experimented with different combinations of datasets through careful selection and different parameters. As a result, we found that a dataset of MBTI types with 550 or more data using Logistic Regression with solver = liblinear and penalty = L1 as its parameters returned the optimal results, with 0.586 exact match ratio and 0.178 hamming loss scores. Although it was not the one with the highest exact match ratio or lowest hamming loss, we believe that this model is the best option as it still includes data with the [N/S] label instead of completely omitting them.

Working with the selected glass-box models following the XAI approach, we could easily understand the decision-making process of each model, as well as how the results came to be. As we added to each classification results summary for each model, it is very clear and easy to interpret how they work. Also, each of the models used are easily adjustable, which allowed us to experiment with different combinations of parameters and datasets.

For future work, we can analyze the results of each selected model and build a glass-box model of our own that combines the best of all three models and still follow the XAI approach. We could also experiment with black box models with an explainer technique, meaning working with more complex models that are not as transparent, interpretable, or simple, but along with a separate explainer as an extra step (Roelenga, 2021). By comparing the results, we could analyze which one produces more accurate yet explainable results overall.

Our experiments are conducted on text posts posted on social media. Further studies can involve predicting MBTI types based on the contents that the person likes, not just what they write, which would give us further insights into what type of content each MBTI type likes in common. With the optimal MBTI prediction model and analysis of relationships between types and traits, we can apply this to social media and online platforms to provide a more personalized online experience. This could be done by customizing the users' profiles and contents with the ones that fit their personality types and by building a recommendation system that recommends other users with other compatible MBTI types.